\documentclass[letterpaper, conference]{IEEEtran}
\IEEEoverridecommandlockouts

\usepackage{cite}
\usepackage{amsmath,amssymb,amsfonts}

\usepackage{graphicx}
\usepackage{textcomp}
\usepackage{xcolor}

\usepackage{amsthm} 

\usepackage{hyperref}
\usepackage{algorithm}
\usepackage{algpseudocode}
\usepackage{dsfont}
\usepackage{array, booktabs}  
\usepackage{tikz}
\usepackage[top=0.75in, left=0.65in, right=0.65in, bottom=0.98in]{geometry}
\usepackage{multirow}

\def\BibTeX{{\rm B\kern-.05em{\sc i\kern-.025em b}\kern-.08em
    T\kern-.1667em\lower.7ex\hbox{E}\kern-.125emX}}

\title{\LARGE \bf Overcoming Overfitting in Reinforcement Learning via Gaussian Process Diffusion Policy}
\begin{document}

\author{
Amornyos Horprasert\textsuperscript{1},
Esa Apriaskar\textsuperscript{2},
Xingyu Liu\textsuperscript{3},
Lanlan Su\textsuperscript{4} 
and Lyudmila S. Mihaylova\textsuperscript{5}
\thanks{$^{1}$ Amornyos Horprasert is with School of Electrical and Electronic Engineering, University of Sheffield, Sheffield, S1 3JD, United Kingdom {\tt\small ahorprasert1@sheffield.ac.uk}}
\thanks{$^{2}$ Esa Apriaskar is with School of Electrical and Electronic Engineering, University of Sheffield, Sheffield, S1 3JD, United Kingdom, and Department of Electrical Engineering, Faculty of Engineering, Universitas Negeri Semarang, Semarang, 50229, Indonesia {\tt\small eapriaskar1@sheffield.ac.uk}}
\thanks{$^{3}$ Xingyu Liu is with School of Electrical and Electronic Engineering, University of Sheffield, Sheffield, S1 3JD, United Kingdom {\tt\small xliu231@sheffield.ac.uk}}
\thanks{$^{4}$ Lanlan Su is with Department of Electrical and Electronic Engineering, University of Manchester, United Kingdom, Manchester, M13 9PL {\tt\small lanlan.su@manchester.ac.uk}}
\thanks{$^{5}$ Lyudmila S. Mihaylova is with School of Electrical and Electronic Engineering, University of Sheffield, Sheffield, S1 3JD, United Kingdom {\tt\small l.s.mihaylova@sheffield.ac.uk}}
}

\maketitle
\begin{abstract}
One of the key challenges that Reinforcement Learning (RL) faces is its limited capability to adapt to a change of data distribution caused by uncertainties. This challenge arises especially in RL systems using deep neural networks as decision makers or policies, which are prone to overfitting after prolonged training on fixed environments. To address this challenge, this paper proposes Gaussian Process Diffusion Policy (GPDP), a new algorithm that integrates diffusion models and Gaussian Process Regression (GPR) to represent the policy. GPR guides diffusion models to generate actions that maximize learned Q-function, resembling the policy improvement in RL. Furthermore, the kernel-based nature of GPR enhances the policy's exploration efficiency under distribution shifts at test time, increasing the chance of discovering new behaviors and mitigating overfitting. Simulation results on the~\href{https://gymnasium.farama.org/environments/mujoco/walker2d/}{Walker2d} benchmark show that our approach outperforms state-of-the-art algorithms under distribution shift condition by achieving around 67.74\% to 123.18\% improvement in the RL's objective function while maintaining comparable performance under normal conditions.
\end{abstract} 
\vspace{1mm}

\section{Introduction}
\label{sec:introduction}
\vspace{-0.5mm}
Reinforcement Learning (RL) \cite{RL_intro_2nd} has been the subject of intensive research over the past several decades. In complex problems, the dynamics of the environment are often unknown, making the modeling of the control system challenging. This is where RL demonstrates its advantages, as it solely relies on signals received from the environment to learn the optimal control strategies. Despite its impressive performance across various control applications as empirically shown in \cite{mnih2013playing, wang2023diffusion, haarnoja2018soft, lillicrap2015continuous}, RL still encounters certain challenges that hinder its effectiveness toward real-world scenarios. One major challenge is its poor adaptability to unseen states at test time, where those states are the results from changing in the environment's dynamics or data distribution. The reason behind this shortcoming is the usage of deep layers of neural networks as the policy in RL. Once trained on fixed training data distributions, the policy network often struggles to adapt the control in different distributions, resulting in an unreliable behavior. This phenomenon exemplifies the overfitting in RL due to distribution shift, as studied in \cite{zhang2018study, nikishin2022primacy, fujimoto2023assessing}.

While the overfitting and distribution shift can be viewed from diverse perspectives, this work primarily focuses on the overfitting that cause a degradation in performance at test time. This type of overfitting arises as a consequence from the distribution shift caused by adversarial attacks or uncertainties similar to the case studied in \cite{zhang2018study, fujimoto2023assessing}. We expect that overcoming this challenge would represent a significant advancement of RL toward real-world environments, where uncertainties are inevitable. For instance, a robot controlled by RL should be capable of recovering its posture after accidental fall, even if it has not been trained to perform that before. 

In this work, we propose a novel RL framework that integrates generative diffusion models with a kernel-based method---Gaussian Process Regression (GPR)---to serve as the policy. We further demonstrate its effectiveness through a case study addressing overfitting under distribution shift. 
This paper begins by providing backgrounds of RL and diffusion models. The application of GPR in RL are then demonstrated in Section \ref{sec:gaussian_guided_diffusion_models} along with an intriguing characteristic that shows potential in mitigating the overfitting. Finally, the proposed approach is evaluated on a \href{https://gymnasium.farama.org/environments/mujoco/walker2d/}{Walker2D} problem from OpenAI Gym \cite{brockman2016openai} followed by a conclusion of findings and limitations. 

\section{Preliminaries}
\label{sec:preliminaries}
\textbf{Reinforcement Learning.} The RL problems are typically formulated as Markov Decision Process (MDP)\cite{RL_intro_2nd} with tuples (\(\mathcal{S},\mathcal{A}, T,r, \gamma\)). 
At each time step \(t\), the agent performs an action \(\mathbf{a}_t\in\mathcal{A}\) based on its current knowledge or policy \(\boldsymbol{\pi}\). The agent then perceives the state \(\mathbf{s}_{t+1}\in\mathcal{S}\) as a feedback. The objective is to acquire a policy that maximizes the expected cumulative discount reward, expressed as:
\begin{align}
    \hspace{-6pt}
    \mathcal{J}(\boldsymbol{\pi}_{\boldsymbol{\theta}}) = \mathbb{E}_{\substack{\mathbf{s}_0\sim d_0(\mathbf{s}_0), \hspace{1mm}\mathbf{a}_t\sim\boldsymbol{\pi}_{\boldsymbol{\theta}}(\mathbf{a}_t|\mathbf{s}_t)\\\mathbf{s}_{t+1}\sim T(\mathbf{s}_{t+1}|\mathbf{s}_t,\mathbf{a}_t)}} \bigg[\sum^{H-1}_{t=0}\hspace{-0.5mm}\gamma^t r(\mathbf{s}_t, \mathbf{a}_t, \mathbf{s}_{t+1})\bigg],
    \label{eq:rl_objective}
\end{align}
where \(T(\mathbf{s}_{t+1}|\mathbf{s}_t,\mathbf{a}_t)\) is the state transition dynamics, \(d_0(\mathbf{s}_0)\) denotes the distribution of initial state, \(r(\mathbf{s}_t, \mathbf{a}_t, \mathbf{s}_{t+1}) \in \mathbb{R}\) is a reward function, and \(\gamma\in[0,1]\) is a discount factor. \(H\) could be infinite but in the case where the interaction is episodic, \(H\) is the trajectory length of the episode. The policy \(\boldsymbol{\pi}_{\boldsymbol{\theta}}(\mathbf{a}_t|\mathbf{s}_t)\) is a function parameterized by \(\boldsymbol{\theta}\). The policy \(\boldsymbol{\pi}\) can also learn from a static dataset \(\mathcal{D}\), which contains MDP transitions \(\mathcal{D} = \{(\mathbf{s}_k, \mathbf{a}_k, r_k, \mathbf{s}_{k+1})\}^{n-1}_{k=0}\) pre-collected by any behavior policy, denoted as \(\boldsymbol{\pi}_b\). This area of RL is called \textit{Offline Reinforcement Learning} (Offline-RL)\cite{levine2020offline, kostrikov2022offline}, where interaction with the environment is prohibited at the training phase. 
The proposed approach is going to be designed as an Offline-RL since it benefits the learning nature of diffusion models and GPR.
\vspace{1mm}

\textbf{Diffusion Models.} Diffusion models, introduced by \cite{sohl2015deep, ho2020denoising}, are the generative models in the form of latent variables. They corrupts a complex, intractable data distribution \(q(\mathbf{x}^0)\) by gradually injecting Gaussian noise according to a variance schedule \(\beta^1, \beta^2, ..., \beta^N\). The process is referred to as a \textit{forward process}, which has a form of \(q(\mathbf{x}^{0:N})=q(\mathbf{x}^0)\prod^N_{i=1}q(\mathbf{x}^i|\mathbf{x}^{i-1})\). To generate samples, the forward chain is reversed, creating another trajectory called \textit{reverse process}, which remains in the same form as \(p_{\boldsymbol{\theta}}(\mathbf{x}^{0:N}) = p(\mathbf{x}^N)\prod^N_{i=1}p_{\boldsymbol{\theta}}(\mathbf{x}^{i-1}|\mathbf{x}^i)\). The training involves optimizing a variational bound on a negative log likelihood \(\mathbb{E}[-\log{p_{\boldsymbol{\theta}}(\mathbf{x}^0)}]\).
\vspace{-3mm}

\textbf{Diffusion Policies.} Diffusion models can be exploited as the policy \(\boldsymbol{\pi}\) in RL through slight modifications to the Markov chain, as proposed by \cite{chi2023diffusion, pearce2023imitating}. The notation for data variable is changed from ``\(\mathbf{x}\)" to ``\(\mathbf{a}\)" to represent an action, and the reverse process is conditioned on the state \(\mathbf{s}\). However, these modifications do not alter the form of the reverse diffusion chain, which is still expressed as:
\vspace{-1mm}
\begin{align}
    \boldsymbol{\pi}_{\boldsymbol{\theta}}(\mathbf{a}_t|\mathbf{s}_t) = p_{\boldsymbol{\theta}}(\mathbf{a}^{0:N}|\mathbf{s}_t) &= p(\mathbf{a}^N)\prod_{i=1}^Np_{\boldsymbol{\theta}}(\mathbf{a}^{i-1}|\mathbf{a}^i,\mathbf{s}_t),
    \label{eq:diffusion_policy}
\end{align}
%
where \(p(\mathbf{a}^N)=\mathcal{N}(\mathbf{a}^N|\mathbf{0},\mathbf{I})\) is a starting point for the chain, and \(\mathbf{I}\) is an identity matrix. A sample at the final step is used for the interaction (\(\text{i.e., }\mathbf{a}_t\sim p_{\boldsymbol{\theta}}(\mathbf{a}^{0:N}|\mathbf{s}_t)\)). The intermediate distribution can be estimated followed \cite{ho2020denoising}, given by:
%
\begin{align}
    p_{\boldsymbol{\theta}}(\mathbf{a}^{i-1}|\mathbf{a}^i, \mathbf{s}_t) = \mathcal{N}(\mathbf{a}^{i-1}|\boldsymbol{\mu}_{\boldsymbol{\theta}}(\mathbf{s}_t, \mathbf{a}^i,i),\boldsymbol{\Sigma}_{\boldsymbol{\theta}}),
    \label{eq:intermediate_reverse_process}
\end{align}
where \(\boldsymbol{\mu}_{\boldsymbol{\theta}}(\mathbf{s}_t,\mathbf{a}^i,i)=\frac{1}{\sqrt{1-\beta}^i}(\mathbf{a}^i-\frac{\beta^i}{\sqrt{1-\bar{\alpha}^i}}\boldsymbol{\epsilon}_{\boldsymbol{\theta}}(\mathbf{s}_t,\mathbf{a}^i,i))\), \(\beta^i\) is a diffusion rate at diffusion step \(i\)
, \(\bar{\alpha}^i=\prod^i_{j=1}(1-\beta^j)\), \(\boldsymbol{\Sigma}_{\boldsymbol{\theta}}=\beta^i\mathbf{I}\), and \(\boldsymbol{\epsilon}_{\boldsymbol{\theta}}(\mathbf{s}_t,\mathbf{a}^i,i)\) is a learned residual noise estimator. The objective function can be simplified from the intractable negative log likelihood into a tractable form:
\begin{align}
    \hspace{-7pt}
    \mathcal{L}_{\boldsymbol{\epsilon}}(\boldsymbol{\theta}) = \mathbb{E}_{(\mathbf{s},\mathbf{a},\boldsymbol{\epsilon},i)}\biggr[ ||\boldsymbol{\epsilon} - \boldsymbol{\epsilon}_{\boldsymbol{\theta}}(\mathbf{s}, \sqrt{\bar{\alpha}^i}\mathbf{a} + \sqrt{1-\bar{\alpha}^i}\boldsymbol{\epsilon}, i) ||^2\biggr],
    \label{eq:diffusion_policy_objective}
\end{align}
where \((\mathbf{s},\mathbf{a})\sim\mathcal{D}\), \(\boldsymbol{\epsilon}\sim\mathcal{N}(\mathbf{0},\mathbf{I})\), and \(i\) is sampled from a uniform distribution over diffusion timestep (\(i \sim U(1,N)\)). The goal of diffusion policy is to imitate or achieve similar performance to the behavior policy (\(\boldsymbol{\pi}_{\boldsymbol{\theta}}\approx\boldsymbol{\pi}_b\)).

\section{Gaussian Process Diffusion Policy}
\label{sec:gaussian_guided_diffusion_models}
\vspace{-0.5mm}
As previously mentioned, the diffusion policy's objective is to merely mimic \(\boldsymbol{\pi}_b\) but we want \(\boldsymbol{\pi}_{\boldsymbol{\theta}}\) to perform better than \(\boldsymbol{\pi}_b\) in terms of maximizing the cumulative reward (i.e., \(\mathcal{J}(\boldsymbol{\pi}_{\boldsymbol{\theta}}) \geq \mathcal{J}(\boldsymbol{\pi}_b)\)). Therefore, it is necessary to evaluate and improve \(\boldsymbol{\pi}_{\boldsymbol{\theta}}\) as in the RL framework. In this section, we present a way to apply Gaussian Process Regression (GPR) into the diffusion policy as the policy improvement. Then we design the policy evaluation process by leveraging an existing algorithm, resulting in a complete RL framework. Finally, we discuss an interesting property related to handling uncertainty, which highlights its potential to mitigate the overfitting challenge mentioned earlier.
\vspace{1mm}

\textbf{Gaussian-Guided Reverse Process.} As proposed in \cite{sohl2015deep}, the reverse process can be modified by multiplying it with another sufficiently smooth distribution, denoted as \(g(\mathbf{y})\). This creates another form of the intermediate distribution that has a perturbation on its mean, given by:
\begin{align}
    \tilde{p}_{\boldsymbol{\theta}}(\mathbf{a}^{i-1}|\mathbf{a}^i,\mathbf{s}_t) &= \mathcal{N}(\mathbf{a}^{i-1}|\boldsymbol{\mu}_{\boldsymbol{\theta}}+\boldsymbol{\Sigma}_{\boldsymbol{\theta}}\mathbf{g},\boldsymbol{\Sigma}_{\boldsymbol{\theta}}), 
    \label{eq:sampling_x_i_1_perturbed}
\end{align}
where \(\mathbf{g}=\frac{\partial\log{g(\mathbf{y})}}{\partial \mathbf{y}}\big|_{\mathbf{y}=\boldsymbol{\mu}_{\boldsymbol{\theta}}}\), and to make the notation cleaner, we abbreviate \(\boldsymbol{\mu}_{\boldsymbol{\theta}}(\mathbf{s}_t,\mathbf{a}_t,i) = \boldsymbol{\mu}_{\boldsymbol{\theta}}\). As a result, the reverse process can be guided towards a specific output by the guidance made from the distribution \(g(\mathbf{y})\). The reverse Markov chain also remains in the same form as \(\tilde{p}_{\boldsymbol{\theta}}(\mathbf{a}^{0:N}|\mathbf{s}_t)=\tilde{p}(\mathbf{a}^N)\prod^N_{i=1}\tilde{p}_{\boldsymbol{\theta}}(\mathbf{a}^{i-1}|\mathbf{a}^i,\mathbf{s}_t)\) and \(\tilde{p}(\mathbf{a}^N)=\mathcal{N}(\mathbf{a}^N|\mathbf{0},\mathbf{I})\). 
\vspace{1mm}

In \cite{sohl2015deep, dhariwal2021diffusion}, \(g(\mathbf{y})\) was exploited as a learned classifier to guide the diffusion model in generating desired images. In this work, the use of \(g(\mathbf{y})\) is adapted to RL problems. \(g_{\boldsymbol{\omega}}(\mathbf{y})\) is assumed to be a Gaussian distribution (i.e., \(g_{\boldsymbol{\omega}}(\mathbf{y}) = \mathcal{N}(\mathbf{y}|\boldsymbol{\mu}_{\boldsymbol{\omega}},\boldsymbol{\Sigma}_{\boldsymbol{\omega}})\)), with sufficient smoothness and parameterized by \(\boldsymbol{\omega}\). Under these assumptions, substituting the density function of \(g_{\boldsymbol{\omega}}(\mathbf{y})\) into (\ref{eq:sampling_x_i_1_perturbed}) allows a closed-form derivation as: 
\begin{align}
    \hspace{-5pt}
    \tilde{p}_{\boldsymbol{\theta}}(\mathbf{a}^{i-1}|\mathbf{a}^{i},\mathbf{s}_t) = \mathcal{N}(\mathbf{a}^{i-1}|\boldsymbol{\mu}_{\boldsymbol{\theta}}-\boldsymbol{\Sigma}_{\boldsymbol{\theta}}\boldsymbol{\Sigma}_{\boldsymbol{\omega}}^{-1}(\boldsymbol{\mu}_{\boldsymbol{\theta}}-\boldsymbol{\mu}_{\boldsymbol{\omega}}), \boldsymbol{\Sigma}_{\boldsymbol{\theta}}).
    \label{eq:guided_distribution}
\end{align}

Here, the perturbation on the mean \(\boldsymbol{\mu}_{\boldsymbol{\theta}}\) of the reverse process is described by the probabilistic properties of \(g_{\boldsymbol{\omega}}(\mathbf{y})\). We construct this perturbed form as the policy (\(\boldsymbol{\pi}_{\boldsymbol{\theta}}(\mathbf{a}^0|\mathbf{s}_t)=\tilde{p}_{\boldsymbol{\theta}}(\mathbf{a}^{0:N}|\mathbf{s}_t)\)). Although, \(g_{\boldsymbol{\omega}}(\mathbf{y})\) can be estimated through various approaches, the GPR method is a suitable choice for this role, as its predictive distribution is inherently Gaussian as mentioned in \cite{10.5555/1162264, Rasmussen2006Gaussian}. Additionally, the kernel-based nature of GPR can enhance the diffusion policy with the uncertainty awareness capability, which is an essential feature that will be discussed in more detail later in this chapter. 
\vspace{1mm}

\textbf{Estimating the Guidance Distribution.} 
Let the trajectory length \(H\) be the number of samples stored in the training matrices. To adapt the notation for RL problems, we define a matrix of training input as a state matrix \(\mathbf{S} = [\mathbf{s}_0, \mathbf{s}_1,...,\mathbf{s}_{H-1}]\), where each state vector has \(d\) dimensions (\(\mathbf{S}\in\mathbb{R}^{H\times d}\)). Similarly, since the GPR is employed to predict actions for RL, the observation matrix is replaced by a training action matrix, denoted as \(\mathbf{A} = [\mathbf{a}_0, \mathbf{a}_1,...,\mathbf{a}_{H-1}]\), where each action vector \(\mathbf{a}\) has \(m\) dimensions (\(\mathbf{A}\in\mathbb{R}^{H\times m}\)). The \(\boldsymbol{\mu}_{\boldsymbol{\omega}}\) and \(\boldsymbol{\Sigma}_{\boldsymbol{\omega}}\) in (\ref{eq:guided_distribution}) can be obtained from (\ref{eq:gp_mean}) and (\ref{eq:gp_cov}), which are probabilistic properties of the distribution over zero-mean function given the noisy action matrix \(\mathbf{A}\) with variance \(\sigma^2_n\) as given by: 
\begin{align}
    \boldsymbol{\mu}_{\boldsymbol{\omega}} &= \mathbf{K}_{\mathbf{S}_*\mathbf{S}}(\mathbf{K}_{\mathbf{S}\mathbf{S}}+\sigma_n^2\mathbf{I})^{-1}\mathbf{A}, 
    \label{eq:gp_mean}\\
    \boldsymbol{\Sigma}_{\boldsymbol{\omega}} &= \mathbf{K}_{\mathbf{S}_*\mathbf{S}_*} - \mathbf{K}_{\mathbf{S}_*\mathbf{S}}(\mathbf{K}_{\mathbf{S}\mathbf{S}}+\sigma_n^2\mathbf{I})^{-1}\mathbf{K}_{\mathbf{S}\mathbf{S}_*}, \label{eq:gp_cov}
\end{align}
\noindent where \(\mathbf{S}_*\) represents a matrix of test input, corresponding to the next observation at time \(t\), i.e., \(\mathbf{S}_* = \mathbf{s}_{t+1}\sim T(\mathbf{s}_{t+1}|\mathbf{s}_t,\mathbf{a}_t)\), \(\mathbf{K}_{\mathbf{S}\mathbf{S}_*} = [k(\mathbf{s}_x,\mathbf{s}_{t+1})]^{H-1}_{x=0} \in \mathbb{R}^{H\times 1}\) denotes the kernel matrix, where \(k(\cdot,\cdot)\) is a kernel function computed at all pair of training and test input point, and similarly for other matrices \(\mathbf{K}_{\mathbf{S}_*\mathbf{S}}=\mathbf{K}_{\mathbf{S}\mathbf{S}_*}^\top, \mathbf{K}_{\mathbf{S}\mathbf{S}}=[k(\mathbf{s}_x,\mathbf{s}_w)]^{H-1}_{x,w=0}\in\mathbb{R}^{H\times H}, \text{ and }\mathbf{K}_{\mathbf{S}_*\mathbf{S}_*}=k(\mathbf{s}_{t+1},\mathbf{s}_{t+1})\in\mathbb{R}\). The kernel function choice is flexible and depended on the problem's complexity. We choose the basic square exponential (SE) kernel as we found that it can provide a decent performance for the experiment carried out in this work. The SE kernel function can be expressed as:
\begin{align}
    k({\mathbf{s}_1,\mathbf{s}_2}) &= \sigma_p^2\exp{(-\frac{1}{2\ell^2}\Vert\mathbf{s}_1-\mathbf{s}_2\Vert^2)},
    \label{eq:se_kernel}
\end{align}
where \(\sigma_p\) and \(\ell\) are hyperparameters, \(\mathbf{s}_1\) and \(\mathbf{s}_2\) denote respectively the entries of input matrices \(\mathbf{S}\) or \(\mathbf{S}_*\), depending on which kernel matrix is being derived. It can be found from the SE kernel's expression (\ref{eq:se_kernel}) that there are two hyperparameters that need to be optimized. Combining them with the observation variance \(\sigma_n^2\), a set of hyperparameters \(\boldsymbol{\omega}\) contains \(\boldsymbol{\omega} = \{\sigma_n, \sigma_p, \ell\}\). The parameter \(\boldsymbol{\omega}\) is optimized by minimizing the negative marginal log-likelihood given by: 
\begin{equation}
    \begin{split}
        \mathcal{L}_g(\boldsymbol{\omega}) = \frac{1}{2}\mathbf{A}^\top(\mathbf{K}_{\mathbf{S}\mathbf{S}}+\sigma^2_n\mathbf{I})^{-1}\mathbf{A} 
        + \frac{1}{2}\log|\mathbf{K}_{\mathbf{S}\mathbf{S}}+\sigma^2_n\mathbf{I}| \hspace{-1mm} \\
        +\frac{H}{2}\log{2\pi}.
    \end{split}
    \label{eq:negative_log_likelihood_original}
\end{equation}

\textbf{Imitating Policy Improvement via Gaussian Processes.} To acquire \(\mathcal{J}(\boldsymbol{\pi}_{\boldsymbol{\theta}})\geq\mathcal{J}(\boldsymbol{\pi}_b)\), the standard GPR method is required to be modified according to the following procedures.

First, the states of best trajectory from the dataset are stored in the training state matrix (\(\mathbf{S} \leftarrow \mathbf{S} \cup \{\mathbf{s}^{\text{best}}_0, \mathbf{s}^{\text{best}}_1, ..., \mathbf{s}^{\text{best}}_{H-1}\}\)). This implies that the MDP transitions stored in \(\mathcal{D}\) must be time-dependent, meaning that the RL problem is episodic. If not, the problem should be formulated such that the states associated with the best reward can be accessed from the initial state \(\mathbf{s}_0\). Otherwise, the agent will be unable to reach those states, resulting in high variance in the predictions from GPR. Additionally, care must be taken regarding the size of \(\mathbf{S}\). It should be limited to a few thousand data points due to the use of exact inference, which possesses the time complexity of \(\mathcal{O}(H^3)\) from the inversion of the kernel matrix. 

Intuitively, after the states of best trajectory are stored, the set of observation \(\mathbf{A}\) should be stored with the actions associated to the best trajectory as well. However, since we seek our policy to perform better than \(\boldsymbol{\pi}_b\), the set of observation will be stored by actions that ``greedily" maximize the expected cumulative reward at each time step in the trajectory, denoted as \(\mathbf{a}^{\text{alt}}_t\) (\(\mathbf{A}\leftarrow\mathbf{A}\cup\{\mathbf{a}^{\text{alt}}_0, \mathbf{a}^{\text{alt}}_1, ..., \mathbf{a}^{\text{alt}}_{H-1}\}\)). The \(\mathbf{a}^{\text{alt}}_t\) can be deterministically sampled from the expression below:
\begin{align}
    \mathbf{a}_t^{\text{alt}} = \underset{\hat{\mathbf{a}}^0_l}{\text{argmax}}\hspace{1mm}Q_{\boldsymbol{\phi}}(\mathbf{s}^{\text{best}}_t,\hat{\mathbf{a}}^0_l),
    \label{eq:altered_observation}
\end{align}
where \(Q_{\boldsymbol{\phi}}(\mathbf{s},\mathbf{a})\) is a learned expected cumulative reward function given state and action or Q-function parameterized by \(\boldsymbol{\phi}\), \(\hat{\mathbf{a}}_l^0\sim p_{\boldsymbol{\theta}}(\mathbf{a}^{0:N}|\mathbf{s}^{\text{best}}_t)\) for {\(l=0,1,...,M-1\)}. The idea behind this process is to sample {\(M\)} candidate actions in a given state \(\mathbf{s}^{\text{best}}_t\) from the diffusion policy without guidance (\ref{eq:diffusion_policy}), then select the action that maximizes learned Q-function. By doing so, we can expect that the GPR will give a distribution of \(\mathbf{a}^{\text{alt}}_t\) with low variance as an output, if the GPR is confident in the assessment on the input (e.g., \(\mathbf{S}_* \subseteq \mathbf{S}\)). This altered observation process resembles sampling actions from the greedy policy in conventional Q-learning algorithms \cite{mnih2013playing, lillicrap2015continuous, fujimoto2018addressingfunctionapproximationerror}.
\vspace{1mm}

\textbf{Policy Evaluation.} According to (\ref{eq:altered_observation}), we need an approach to learn \(Q_{\boldsymbol{\phi}}\). A challenge is that the diffusion policy is not explicitly optimized, but rather implicitly improved via the GPR. This raises a problem since most Q-learning algorithms rely on \textit{bootstrapping} method, where the policy network is required to predict \(\mathbf{a}_{t+1}\) during the training of Q-function. However, Implicit Q-learning (IQL), proposed by \cite{kostrikov2022offline} is a suitable choice for this role. The IQL enables the learning of Q-function in an offline manner without the need of bootstrapping to approximate true Q-functions. An additional benefit of using the IQL is that it avoids evaluating out-of-distribution actions (OOD) caused by bootstrapping methods, which is another challenge in Offline-RL. The Q-function \(Q_{\boldsymbol{\phi}}(\mathbf{s}_t,\mathbf{a}_t)\), learns through minimizing the IQL's objective expressed as:
\begin{align}
    \mathcal{L}_Q(\boldsymbol{\phi}) = \mathbb{E}_{(\mathbf{s}_t,\mathbf{a}_t, \mathbf{s}_{t+1})\sim\mathcal{D}}[(\overline{Q}(\mathbf{s}_t,\mathbf{a}_t)-Q_{\boldsymbol{\phi}}(\mathbf{s}_t,\mathbf{a}_t))^2], 
    \label{eq:iql_q_loss_function}
\end{align}
where \(\overline{Q}(\mathbf{s}_t,\mathbf{a}_t)=r(\mathbf{s}_{t},\mathbf{a}_t, \mathbf{s}_{t+1})+\gamma V_{\boldsymbol{\psi}}(\mathbf{s}_{t+1})\), and \(V_{\boldsymbol{\psi}}(\mathbf{s}_{t+1})\) is a value function given the next state \(\mathbf{s}_{t+1}\) parameterized by \(\boldsymbol{\psi}\), which is learned via minimizing the expectile regression on the temporal difference (TD) error, expressed as:
\begin{align}
    \mathcal{L}_V(\boldsymbol{\psi}) = \mathbb{E}_{(\mathbf{s}_t,\mathbf{a}_t)\sim\mathcal{D}}[L_2^T(Q_{\hat{\boldsymbol{\phi}}}(\mathbf{s}_t,\mathbf{a}_t)-V_{\boldsymbol{\psi}}(\mathbf{s}_t))], 
    \label{eq:iql_v_loss_function}
\end{align}
where \(L_2^T(u)=|\tau-\mathds{1}(u<0)|u^2\), \(u=Q_{\hat{\boldsymbol{\phi}}}(\mathbf{s}_t,\mathbf{a}_t)-V_{\boldsymbol{\psi}}(\mathbf{s}_t)\), \(\tau\in(0,1)\) is an expectile of \(u\) or TD error, and \(\hat{\boldsymbol{\phi}}\) is a set of parameters of a target Q-function network, which is updated using soft method as in \cite{lillicrap2015continuous, haarnoja2018soft, kostrikov2022offline}.
\vspace{1mm}

\textbf{Enhancing Policy's Exploration Capabilities under Distribution Shifts.} Due to GPR's ability to quantify uncertainty, the predictive mean and covariance can be varied according to the correlation between \(\mathbf{S}_*\) and \(\mathbf{S}\).
An interesting situation arises when uncertainty causes a shift in the transition dynamics \(T\), leading to a new distribution, denoted as \(T_U(\mathbf{s}_{t+1}|\mathbf{s}_t,\mathbf{a}_t)\). 
Given that \(\mathbf{S}_* = \mathbf{s}_{t+1}\sim T_U(\mathbf{s}_{t+1}|\mathbf{s}_t,\mathbf{a}_t)\), the correlation between \(\mathbf{S}_*\) and \(\mathbf{S}\) can be significantly low. In the case, where the correlation is minimal {(i.e., \(\Vert\mathbf{S}_*-\mathbf{S}\Vert^2\rightarrow\mathbf{\infty}\))}, \(\boldsymbol{\mu}_{\boldsymbol{\omega}}\) and 
\(\boldsymbol{\Sigma}_{\boldsymbol{\omega}}\) approximately become \(\mathbf{0}\) and \(\mathbf{K}_{\mathbf{S}_*\mathbf{S}_*}\) respectively.
Substituting these values into (\ref{eq:guided_distribution}) changes the expression to:
\begin{align}
    \tilde{p}_{\boldsymbol{\theta}}(\mathbf{a}^{i-1}|\mathbf{a}^i,\mathbf{s}_t) \approx \mathcal{N}(\mathbf{a}^{i-1}|\boldsymbol{\mu}_{\boldsymbol{\theta}},\boldsymbol{\Sigma}_{\boldsymbol{\theta}}) = p_{\boldsymbol{\theta}}(\mathbf{a}^{i-1}|\mathbf{a}^i,\mathbf{s}_t).
    \label{eq:guided_reverse_process_in_uncertainty}
\end{align}

Notably, the perturbation term in the mean of the guided reverse process is vanished. As a result, the guided form reverts to the reverse process without guidance as (\ref{eq:intermediate_reverse_process}). This situation allows the diffusion policy to sample from the full range of possible actions in the uncertain state \(\mathbf{s}_{t+1}\), rather than greedy actions sampled from (\ref{eq:altered_observation}). Consequently, this behavior can lead the agent to explore a new set of actions that may yield a better reward under the shifted transition dynamics.

It is important to note that \(\mathbf{s}_{t+1}\) should be treated as an anomaly with respect to the GPR, but must be presented in the main dataset 
(\(\{\mathbf{s}_{t+1}\sim T_U(\mathbf{s}_{t+1}|\mathbf{s}_t,\mathbf{a}_t)\}\in\mathcal{D}\)). Otherwise, the diffusion policy may provide an unreliable output, as the model has not seen this state during training.

In summary, the diffusion policy achieves a better policy than \(\boldsymbol{\pi}_b\) by incorporating the guidance made from GPR, which predicts actions that greedily maximize learned Q-function, resembling the policy improvement. Additionally, the GPR converts the guided diffusion policy to non-guided form, as in (\ref{eq:guided_reverse_process_in_uncertainty}) when encountering novel situations caused by distribution shifts. By the combination of these components, we refer to this framework as Gaussian Process Diffusion Policy (GPDP). 

\section{Evaluations \& Results}
\label{sec:evaluations}
The proposed algorithm is evaluated on the \href{https://gymnasium.farama.org/environments/mujoco/walker2d/}{Walker2d} problem,
which is a bipedal robot with three controllable joints on each leg (\(\mathbf{a}_t\in\mathbb{R}^{1\times6}\)). The motions of all joints are observed as a state \(\mathbf{s}_t\in\mathbb{R}^{1\times17}\), which have dynamics followed the kinematic of the robot (\(T(\mathbf{s}_{t+1}|\mathbf{s}_t,\mathbf{a}_t)\)).
For baseline comparison, we implement Soft Actor-Critic (SAC), proposed by \cite{haarnoja2018soft_original, haarnoja2018soft}. 
A trained SAC agent utilizing a stochastic policy (SAC-S), is employed to generate a dataset \(\mathcal{D}\) with \(n\approx10^6\). The dataset contains an equal amount of sample from expert behaviors (fully completed the episode) and medium-performance behaviors (early terminated the episode). This configuration of \(\mathcal{D}\) is inspired by how the datasets are constructed in the D4RL \cite{fu2020d4rl}, a well-known Offline-RL benchmark.  

A Multi-layer Perceptron (MLP) with three hidden layers, followed by Mish activation function \cite{Misra2020MishAS} is served as the architecture for function approximators \(\boldsymbol{\theta}, \boldsymbol{\phi}, \hat{\boldsymbol{\phi}},\boldsymbol{\psi}\). The expectile value \(\tau=0.7\), with a soft update parameter \(\eta=0.005\), and discount factor \(\gamma=0.99\). All networks are trained using Adam optimizer \cite{kingma2014adam} with a learning rate of \(3\times10^{-4}\) for approximately 2 million steps (512 epochs with a mini-batch size of 256). For the diffusion schedule, we adopt a variance preserving stochastic differential equation (VP-SDE) as proposed in \cite{song2021scorebased, xiao2022tackling}, with \(\beta^{\text{min}}=0.1\) and \(\beta^{\text{max}}=10\) and diffusion step \(N=5\). 
The source code and further implementation details are available online
\footnote{\url{https://github.com/AmornyosH/GPDP_IEEE_SSP_2025}\label{source_code_link}}.
\vspace{1mm}

\textbf{Performance Evaluation.} Apart from the baseline (SAC-S), we also compare the performance of GPDP with certain state-of-the-art algorithms with respect to the non-discounted cumulative reward derived from (\ref{eq:rl_objective}) without the expectation and \(\gamma=1\). The first method is SAC-D, which is derived from SAC-S but the stochasticity is removed at test time \cite{haarnoja2018soft}. Another method is Diffusion-QL, denoted as D-QL \cite{wang2023diffusion}, which demonstrates superior performance among diffusion-based RL algorithms.  
Each algorithm is evaluated over 10 Monte Carlo runs with different seeds. Results on normal situation (no distribution shift) are quoted in the ``Normal" test condition rows in Table~\ref{tab:walker2d_results}. 
While GPDP is slightly inferior to SAC-D and D-QL in average reward, it outperforms SAC-S (baseline) in both average and maximum by achieving 17.19\% and 4.31\% higher results respectively. 
\begin{table}[!t]
\caption{
\textbf{Non-discounted Cumulative Reward on the \href{https://gymnasium.farama.org/environments/mujoco/walker2d/}{Walker2d}.} 
The results are averaged over 10 Monte Carlo runs. The maximum results are quoted from the runs that get highest reward.
}
\begin{center}
    \begin{tabular}{>{\centering}m{1.25cm}  >{\centering}m{1.25cm} >{\centering}m{0.85cm}  >{\centering}m{0.85cm}  >{\centering}m{0.85cm}  >{\centering\arraybackslash}m{0.85cm}}
        \toprule
        \multirow{2}{1.25cm}{\centering \textbf{Test Condition}} & \multirow{2}{1.25cm}{\centering \textbf{Reward Condition}} & \multicolumn{4}{c}{\textbf{Non-Discounted Cumulative Reward}} \\
         &  & SAC-S & SAC-D & D-QL & GPDP \\
        \midrule
        \multirow{2}{1.25cm}{\centering Normal} & Average & 4580.59 & \textbf{5367.99} & 5325.85 & 5301.57 \\
        & Maximum & 5170.57 & \textbf{5367.99} & 5362.50 & \textbf{5367.99} \\
        \midrule
        Distribution & Average & 1899.57 & 1982.94 & 1763.53 & \textbf{2357.63} \\
        Shift & Maximum & 2518.87 & 1982.94 & 1893.16 & \textbf{4225.14} \\
        \bottomrule
    \end{tabular}
\end{center}
\label{tab:walker2d_results}
\vspace{-5mm}
\end{table}
\vspace{1mm}

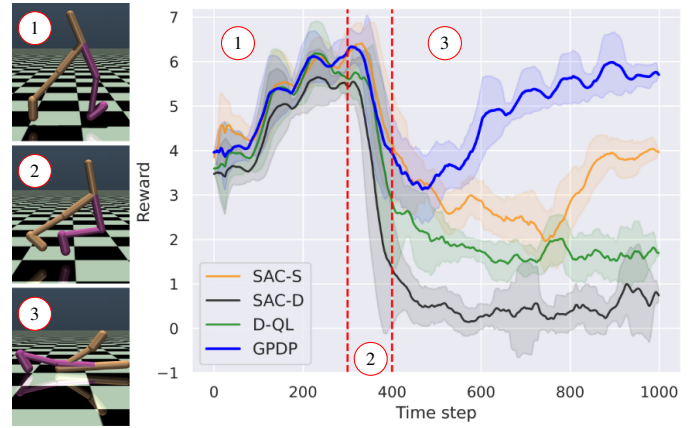
\begin{figure}[!b]
    \centering
    \vspace{-4.0mm}
    \begin{tikzpicture}
        [roundnode/.style={circle, draw=red!100, fill=white!100, minimum size=1mm}]
        \node[anchor=south west,inner sep=0] at (0,0) {\includegraphics[width=1.00\linewidth]{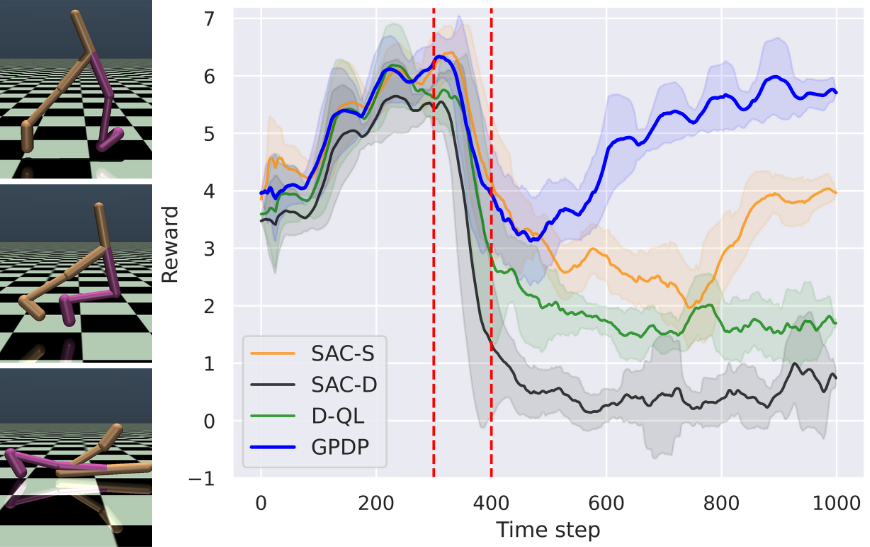}};
        \node[roundnode, scale=0.7] at (0.3,5.3) (first_event) {1};
        \node[roundnode, scale=0.7] at (0.3,3.4) (second_event) {2};
        \node[roundnode, scale=0.7] at (0.3,1.5) (third_event) {3};
        \node[roundnode, scale=0.7] at (3.0,5.1) (first_event) {1};
        \node[roundnode, scale=0.7] at (4.765,0.9) (second_event) {2};
        \node[roundnode, scale=0.7] at (5.75,5.1) (third_event) {3};
    \end{tikzpicture}
    \vspace{-5.0mm}
    \caption{\textbf{Illustration of Distribution Shift.} The left side shows screenshots from \href{https://gymnasium.farama.org/environments/mujoco/walker2d/}{Walker2d}, while the right side presents immediate reward over time in a single trajectory under shifted condition.{\large\textcircled{\small 1}} represents the interval of normal situation. {\large\textcircled{\small 2}} marks the moment when the uncertainty is introduced to the robot. Lastly, {\large\textcircled{\small 3}} indicates when the distribution shift has fully occurred.}
    \label{fig:walker2d_uncertain}
\end{figure}

\textbf{Simulating Distribution Shift.} To emulate the distribution shift caused by environmental uncertainty, all joints in one leg of the robot are disabled after it has been operating for a certain period.  
This disruption persists long enough to ensure that the robot falls to the ground. Once the functionality of the leg is restored, the robot must find a way to regain its reward from a new dynamics (i.e., \(T\) is shifted to \(T_U\)). This scenario is illustrated in Fig. \ref{fig:walker2d_uncertain}, and the results are presented in the ``Distribution Shift" rows in Table \ref{tab:walker2d_results}. The GPDP surpasses other algorithms in both reward conditions especially in maximum, where it obtains around 67.74\% to 123.18\% improvement in regaining the cumulative rewards.


\section{Conclusion}
\label{sec:conclusion}
This work introduces GPDP, a novel RL framework that integrates a diffusion policy with Gaussian Process Regression (GPR) to serve as the policy.
The performance of GPDP is demonstrated in the Section \ref{sec:evaluations}, where it outperforms state-of-the-art algorithms especially in the distribution shift condition. The reported results imply that the exploration capability of GPDP elaborated at the end of Section \ref{sec:gaussian_guided_diffusion_models}, are capable of discovering new set of actions under unseen states, mitigating the overfitting problem as expected. However, GPDP still possesses several challenges. For instance, the limited sample size in GPR may constrain the overall performance of GPDP, as only a small portion of the dataset is utilized for GPR's training. Another challenge lies in the stochasticity of the policy. The results under distribution shift reveal a significant margin between average and maximum score, suggesting inconsistent performance across multiple runs. 

\newpage
\section*{Acknowledgement}
\label{acknowledgment}
We are grateful to the Office of the Civil Service Commission of Thailand for funding the PhD research of Amornyos Horprasert, to the UK EPSRC through Project NSF-EPSRC: ShiRAS. Towards Safe and Reliable Autonomy in Sensor Driven Systems, under Grant EP/T013265/1, and by the USA National Science Foundation under Grant NSF ECCS 1903466. This work was also supported by the UKRI Trustworthy Autonomous Systems Node in Resilience (REASON) EP/V026747/1 project.

\label{references}
\bibliographystyle{ieeetr} 
\bibliography{references} 

\begin{thebibliography}{10}

\bibitem{RL_intro_2nd}
R.~S. Sutton and A.~G. Barto, {\em {Reinforcement} {Learning}: {An} {Introduction}}.
\newblock Cambridge, {MA}: The MIT Press, second edition~ed., 2018.

\bibitem{mnih2013playing}
V.~Mnih, K.~Kavukcuoglu, D.~Silver, A.~Graves, I.~Antonoglou, D.~Wierstra, and M.~Riedmiller, ``Playing {Atari} with {Deep} {Reinforcement} {Learning},'' {\em arXiv preprint arXiv:1312.5602}, 2013.

\bibitem{wang2023diffusion}
Z.~Wang, J.~J. Hunt, and M.~Zhou, ``Diffusion {Policies} as an {Expressive} {Policy} {Class} for {Offline} {Reinforcement} {Learning},'' in {\em Proceedings of the 11th International Conference on Learning Representations}, 2023.

\bibitem{haarnoja2018soft}
T.~Haarnoja, A.~Zhou, K.~Hartikainen, G.~Tucker, S.~Ha, J.~Tan, V.~Kumar, H.~Zhu, A.~Gupta, P.~Abbeel, {\em et~al.}, ``Soft {Actor-Critic} {Algorithms} and {Applications},'' {\em arXiv preprint arXiv:1812.05905}, 2018.

\bibitem{lillicrap2015continuous}
T.~P. Lillicrap, J.~J. Hunt, A.~Pritzel, N.~Heess, T.~Erez, Y.~Tassa, D.~Silver, and D.~Wierstra, ``Continuous {Control} {With} {Deep} {Reinforcement} {Learning},'' {\em arXiv preprint arXiv:1509.02971}, 2015.

\bibitem{zhang2018study}
C.~Zhang, O.~Vinyals, R.~Munos, and S.~Bengio, ``A {Study} on {Overfitting} in {Deep} {Reinforcement} {Learning},'' {\em arXiv preprint arXiv:1804.06893}, 2018.

\bibitem{nikishin2022primacy}
E.~Nikishin, M.~Schwarzer, P.~D’Oro, P.-L. Bacon, and A.~Courville, ``The {Primacy} {Bias} in {Deep} {Reinforcement} {Learning},'' in {\em Proceedings of International Conference on Machine Learning}, pp.~16828--16847, PMLR, 2022.

\bibitem{fujimoto2023assessing}
T.~Fujimoto, J.~Suetterlein, S.~Chatterjee, and A.~Ganguly, ``Assessing the {Impact} of {Distribution} {Shift} on {Reinforcement} {Learning} {Performance},'' in {\em Proceedings of NeurIPS 2023 Workshop on Regulatable ML}, 2023.

\bibitem{brockman2016openai}
G.~Brockman, ``{OpenAI} {Gym},'' {\em arXiv preprint arXiv:1606.01540}, 2016.

\bibitem{levine2020offline}
S.~Levine, A.~Kumar, G.~Tucker, and J.~Fu, ``Offline {Reinforcement} {Learning}: {Tutorial}, {Review}, and {Perspectives} on {Open} {Problems},'' {\em arXiv preprint arXiv:2005.01643}, 2020.

\bibitem{kostrikov2022offline}
I.~Kostrikov, A.~Nair, and S.~Levine, ``Offline {Reinforcement} {Learning} with {Implicit} {Q-Learning},'' in {\em Proceedings of International Conference on Learning Representations}, 2022.

\bibitem{sohl2015deep}
J.~Sohl-Dickstein, E.~Weiss, N.~Maheswaranathan, and S.~Ganguli, ``Deep {Unsupervised} {Learning} {Using} {Nonequilibrium} {Thermodynamics},'' in {\em Proceedings of the 32nd International Conference on Machine Learning}, pp.~2256--2265, PMLR, 2015.

\bibitem{ho2020denoising}
J.~Ho, A.~Jain, and P.~Abbeel, ``Denoising {Diffusion} {Probabilistic} {Models},'' in {\em Proceedings of Advances in Neural Information Processing Systems} (H.~Larochelle, M.~Ranzato, R.~Hadsell, M.~Balcan, and H.~Lin, eds.), vol.~33, pp.~6840--6851, Curran Associates, Inc., 2020.

\bibitem{chi2023diffusion}
C.~Chi, Z.~Xu, S.~Feng, E.~Cousineau, Y.~Du, B.~Burchfiel, R.~Tedrake, and S.~Song, ``Diffusion {Policy}: Visuomotor {Policy} {Learning} via {Action} {Diffusion},'' {\em The International Journal of Robotics Research}, p.~02783649241273668, 2023.

\bibitem{pearce2023imitating}
T.~Pearce, T.~Rashid, A.~Kanervisto, D.~Bignell, M.~Sun, R.~Georgescu, S.~V. Macua, S.~Z. Tan, I.~Momennejad, K.~Hofmann, and S.~Devlin, ``Imitating {Human} {Behaviour} with {Diffusion} {Models},'' in {\em Proceedings of the 11th International Conference on Learning Representations}, 2023.

\bibitem{dhariwal2021diffusion}
P.~Dhariwal and A.~Q. Nichol, ``Diffusion {Models} {Beat} {GAN}s on {Image} {Synthesis},'' in {\em Proceedings of Advances in Neural Information Processing Systems} (A.~Beygelzimer, Y.~Dauphin, P.~Liang, and J.~W. Vaughan, eds.), 2021.

\bibitem{10.5555/1162264}
C.~M. Bishop, {\em Pattern Recognition and Machine Learning (Information Science and Statistics)}.
\newblock Berlin, Heidelberg: Springer-Verlag, 2006.

\bibitem{Rasmussen2006Gaussian}
C.~E. Rasmussen and C.~K.~I. Williams, {\em Gaussian {Processes} for {Machine} {Learning}}.
\newblock The MIT Press, 2006.

\bibitem{fujimoto2018addressingfunctionapproximationerror}
S.~Fujimoto, H.~Hoof, and D.~Meger, ``Addressing {Function} {Approximation} {Error} {In} {Actor-Critic} {Methods},'' in {\em Proceedings of the 35th International Conference on Machine Learning}, pp.~1587--1596, PMLR, 2018.

\bibitem{haarnoja2018soft_original}
T.~Haarnoja, A.~Zhou, P.~Abbeel, and S.~Levine, ``Soft {Actor-Critic}: {Off-Policy} {Maximum} {Entropy} {Deep} {Reinforcement} {Learning} with a {Stochastic} {Actor},'' in {\em Proceedings International Conference on Machine Learning}, pp.~1861--1870, PMLR, 2018.

\bibitem{fu2020d4rl}
J.~Fu, A.~Kumar, O.~Nachum, G.~Tucker, and S.~Levine, ``{D4RL}: {Datasets} {For} {Deep} {Data-Driven} {Reinforcement} {Learning},'' {\em arXiv preprint arXiv:2004.07219}, 2020.

\bibitem{Misra2020MishAS}
D.~Misra, ``Mish: A {Self} {Regularized} {Non-Monotonic} {Activation} {Function},'' in {\em Proceedings of British Machine Vision Conference}, 2020.

\bibitem{kingma2014adam}
D.~P. Kingma and J.~Ba, ``Adam: A method for stochastic optimization,'' {\em arXiv preprint arXiv:1412.6980}, 2014.

\bibitem{song2021scorebased}
Y.~Song, J.~Sohl-Dickstein, D.~P. Kingma, A.~Kumar, S.~Ermon, and B.~Poole, ``{Score-Based} {Generative} {Modeling} through {Stochastic} {Differential} {Equations},'' in {\em Proceedings of International Conference on Learning Representations}, 2021.

\bibitem{xiao2022tackling}
Z.~Xiao, K.~Kreis, and A.~Vahdat, ``Tackling the {Generative} {Learning} {Trilemma} with {Denoising} {Diffusion} {GAN}s,'' in {\em Proceedings of International Conference on Learning Representations}, 2022.

\end{thebibliography}
    
\end{document}